\documentclass{article}

\usepackage[preprint]{neurips_2025}
\usepackage[utf8]{inputenc} 
\usepackage[T1]{fontenc}    
\usepackage{hyperref}       
\usepackage{url}            
\usepackage{booktabs}       
\usepackage{amsfonts}       
\usepackage{microtype}      
\usepackage{amsmath, amssymb, amsthm}
\usepackage{multirow}
\usepackage{array}          
\usepackage{graphicx}
\usepackage{siunitx} 
\usepackage{multirow}
\usepackage{caption}

\title{Benchmarking GNNs for OOD Materials Property Prediction with Uncertainty Quantification}

\author{%
  Liqin Tan \\
  School of Computer Science and Engineering\\
  Sun Yat-sen University\\
  Guangdong, China\\
  \texttt{tanlq8@mail2.sysu.edu.cn} \\
  \And
  Pin Chen \\
  School of Computer Science and Engineering\\
  Sun Yat-sen University\\
  Guangdong, China\\
  \texttt{chenp85@mail.sysu.edu.cn} \\
  \And
  Menghan Liu \\
  School of Computer Science and Engineering\\
  Sun Yat-sen University\\
  Guangdong, China\\
  \texttt{liumh59@mail2.sysu.edu.cn} \\
  \And
  Xiean Wang \\
  School of Computer Science and Engineering\\
  Sun Yat-sen University\\
  Guangdong, China\\
  \texttt{wangxan@mail2.sysu.edu.cn} \\
  \AND
  Jianhuan Cen \\
  School of Computer Science and Engineering\\
  Sun Yat-sen University\\
  Guangdong, China\\
  \texttt{cenjh3@mail2.sysu.edu.cn} \\
  \And
  Qingsong Zou\thanks{Corresponding author.} \\
  School of Computer Science and Engineering\\
  Sun Yat-sen University\\
  Guangdong, China\\
  \texttt{mcszqs@mail.sysu.edu.cn} \\
}

\begin{document}

\maketitle

\begin{abstract}
    We present MatUQ, a benchmark framework for evaluating graph neural networks (GNNs) on out-of-distribution (OOD) materials property prediction with uncertainty quantification (UQ). MatUQ comprises 1,375 OOD prediction tasks constructed from six materials datasets using five OFM-based and a newly proposed structure-aware splitting strategy, SOAP-LOCO, which captures local atomic environments more effectively. We evaluate 12 representative GNN models under a unified uncertainty-aware training protocol that combines Monte Carlo Dropout and Deep Evidential Regression (DER), and introduce a novel uncertainty metric, D-EviU, which shows the strongest correlation with prediction errors in most tasks. Our experiments yield two key findings. First, the uncertainty-aware training approach significantly improves model prediction accuracy, reducing errors by an average of 70.6\% across challenging OOD scenarios. Second, the benchmark reveals that no single model dominates universally: earlier models such as SchNet and ALIGNN remain competitive, while newer models like CrystalFramer and SODNet demonstrate superior performance on specific material properties. These results provide practical insights for selecting reliable models under distribution shifts in materials discovery.
\end{abstract}

\section{Introduction}
\label{sec:Introduction}

Accurate prediction of material properties is crucial in materials informatics, enabling applications like high-throughput screening, inverse design, and structure optimization \cite{butler2018machine, choudhary2022recent, reiser2022graph, wines2023inverse, merchant2023scaling, zhao2023physics}. Graph neural networks (GNNs) \cite{xie2018crystal, omee2022scalable} have emerged as a leading approach due to their ability to model atomic graphs, capturing relational and spatial information more effectively than traditional descriptor-based methods \cite{MatbenchDunn2020, fung2021benchmarking}.

However, GNNs face two key challenges in real-world materials discovery. First, they often struggle with \textit{out-of-distribution (OOD)} generalization, meaning they fail to accurately predict properties of novel materials that differ significantly from the training data.
 Standard evaluations using random splits overestimate performance by ignoring redundancies in materials datasets \cite{SparseXY}, while real-world tasks require extrapolation to structurally distinct compounds.
 Second, GNNs typically lack robust \textit{uncertainty quantification (UQ)}, which is critical for assessing prediction reliability, especially in data-sparse OOD scenarios where overconfident errors can mislead downstream decisions.
 Although prior work has addressed OOD robustness \cite{SparseXY,NEURIPS2020_f5496252} or UQ \cite{li2024uncertainty, uq2023materials} independently, a unified benchmark evaluating both in a realistic setting is missing.

To address these gaps, we propose MatUQ, a unified benchmark framework for evaluating GNNs under distribution shifts with a focus on uncertainty. We construct a suite of 1,375 OOD prediction tasks from six materials property datasets, covering electronic, thermodynamic, and mechanical properties. We train twelve representative GNN models using an uncertainty-aware strategy that combines Monte Carlo dropout (MCD) and Deep Evidential Regression (DER), and systematically assess their performance using two complementary evaluation dimensions: \textit{predictive accuracy} and \textit{uncertainty quality}.

MatUQ contributes three methodological advances. First, we introduce SOAP-LOCO, a novel structure-based data-splitting strategy utilizing Smooth Overlap of Atomic Positions (SOAP) descriptors \cite{bartok2013representing}. 
This method captures localized atomic environments with high fidelity, enabling rigorous OOD evaluation across five of six datasets.
Second, we develop an uncertainty-aware training protocol
integrating DER with MCD, achieving up to 70.6\% reduction in mean absolute error (MAE) on three datasets while estimating both epistemic and aleatoric uncertainty.
Third, we introduce a refined UQ metric, dropout-enhanced evidential uncertainty (D-EviU), which combines stochastic forward passes with evidential distribution parameters. D-EviU shows superior correlation with prediction errors in most datasets, providing a robust tool for uncertainty evaluation.

Our benchmark reveals key insights. No single GNN architecture performs best across all OOD tasks, 
highlighting the need for task-specific model selection. Earlier models such as SchNet and ALIGNN remain competitive, while newer architectures like CrystalFramer and SODNet excel on specific material properties.
Models with richer geometric priors, such as dynamic frames, bond-angle encoding, or SE(3) equivariance, generally offer better generalization and uncertainty calibration.
Dataset characteristics, including structural diversity, property complexity, and quantum mechanical effects, significantly impact performance, especially for tasks like superconductivity and band gap prediction.

\section{Related work}
\label{sec:Related work}

This work builds upon and integrates progress in three key areas: benchmark design for materials property prediction, generation of OOD evaluation tasks, and UQ methods in deep learning. Below, we review relevant efforts and highlight the gaps that motivate our MatUQ framework.

\subsection{Materials property prediction benchmarks}

Benchmarking has played a critical role in advancing machine learning for materials science. The Matbench suite~\cite{MatbenchDunn2020} established a standardized set of prediction tasks covering diverse properties with unified data splits and metrics. Fung et al.~\cite{fung2021benchmarking} evaluated a wide range of GNN architectures across multiple datasets but focused exclusively on accuracy without considering distribution shifts. Omee et al.~\cite{SparseXY} introduced an important step forward by proposing structure-based OOD splits and demonstrating that GNNs exhibit significantly degraded performance under distribution shifts. However, even their evaluation framework lacks the incorporation of UQ, which is crucial for real-world deployment.

In parallel, domain-specific benchmarks such as JARVIS-DFT~\cite{Jarvis2020} and QMOF~\cite{QMOF2021} offer specialized testbeds with high-fidelity simulation data, but these are often limited in scope or focus on specific classes of materials.
Recently, Varivoda et al. \cite{uq2023materials} benchmarked uncertainty quantification methods for materials property prediction, but without considering distributional shifts in test design. Most importantly, none of the existing benchmarks jointly evaluate the OOD accuracy and uncertainty. MatUQ fills this gap by systematically benchmarking both predictive accuracy and uncertainty quality in 1,375 OOD tasks derived from six material datasets.

\subsection{OOD target generation methods}

To enable OOD evaluation, a critical step is the generation of meaningful data splits that simulate realistic distribution shifts. Prior work commonly applies clustering-based methods using descriptors such as the OFM~\cite{LOCO, SparseXY}. Strategies such as Leave-One-Cluster-Out (LOCO), SparseX, and SparseY create OOD test sets based on sparsely populated regions in composition or property space, enabling model evaluation on low-density regions of the data manifold.

However, these OFM-based strategies rely on global structural or compositional features and may not adequately capture local atomic environments that are essential for fine-grained generalization. As a result, models evaluated under such splits may still benefit from latent structural similarities between training and test data. To address this, we propose SOAP-LOCO, a new structure-based splitting method that leverages SOAP descriptors \cite{bartok2013representing, de2016comparing} to cluster materials. In contrast to global descriptors, SOAP encodes fine-grained atomic environments, which directly affect GNN message passing. 
This approach provides a more realistic and challenging OOD evaluation setting, particularly for GNNs whose predictive capacity heavily depends on atomic-scale structural patterns.

\subsection{Uncertainty quantification methods}

UQ has become a central concern in machine learning applications where safety and reliability are paramount. In materials science, UQ is increasingly recognized as a key requirement for guiding experimental validation and discovery~\cite{li2024uncertainty, uq2023materials}. Broadly, UQ methods fall into several categories: Bayesian neural networks (BNNs)~\cite{BayesianNN1992, dropout2016}, ensemble-based approaches~\cite{ensemble2000, deepensembles2017}, evidential learning~\cite{DeepEvidentialRegression}, and Gaussian process approximations~\cite{gaussian2017}. Each method offers trade-offs between calibration quality and computational cost.

Ensemble methods provide strong performance but incur high computational overhead, while Bayesian approaches offer principled uncertainty estimation but scale poorly to large datasets. DER~\cite{DeepEvidentialRegression} offers a lightweight alternative by estimating the parameters of a predictive distribution in a single forward pass. Dropout-based methods~\cite{dropout2016, MCdropout2017uncertainties, dMC2024} approximate Bayesian inference through stochastic regularization and have been widely applied in both vision and scientific modeling.

Despite this progress, few prior works have evaluated UQ methods jointly under OOD shifts in realistic materials prediction settings. Our work addresses this gap by integrating DER and MCD into a unified training and inference protocol that achieves both accurate and well-calibrated predictions. We also introduce novel evaluation metrics that better reflect the correlation between predictive uncertainty and actual error, enabling more informative model selection and deployment decisions.

\section{MatUQ: a unified benchmark framework for OOD prediction with UQ}
\label{sec:matuq}
\begin{figure}[h!]
    \centering
    \includegraphics[width=\linewidth]{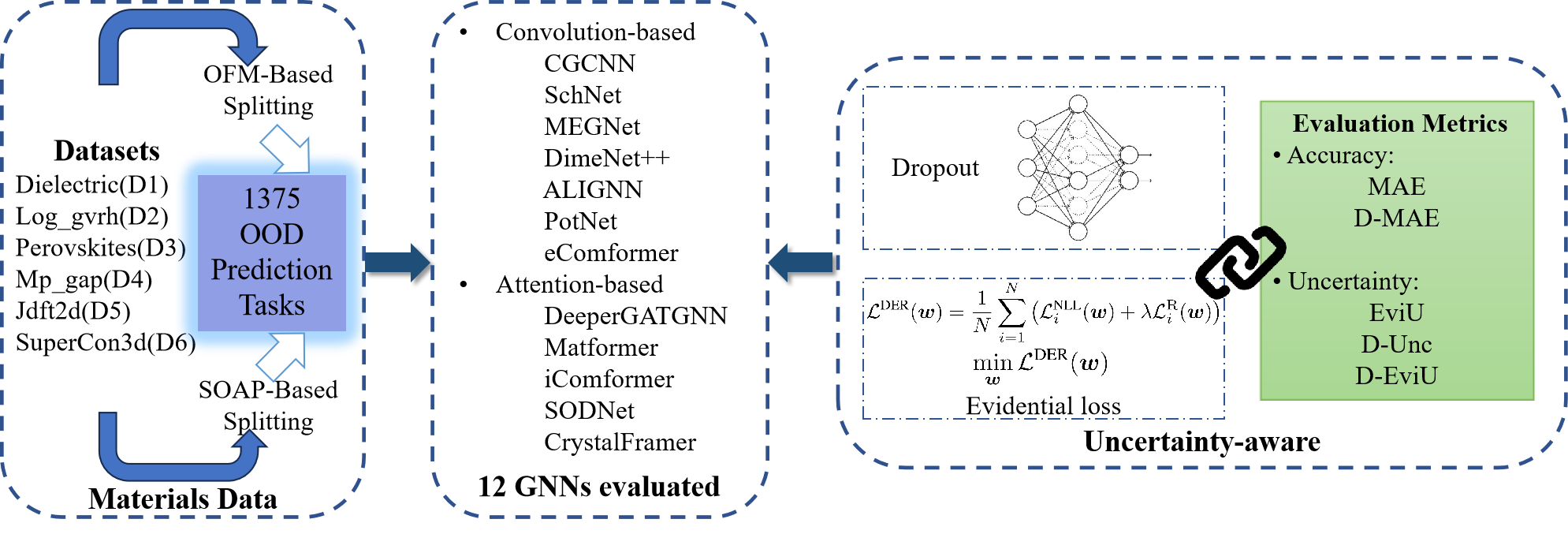}
    \caption{Overview of the MatUQ benchmark framework.}
    \label{fig:overview}
\end{figure}

In this section, we introduce \textbf{MatUQ}, the benchmark framework for evaluating GNNs in the prediction of the properties of OOD materials with UQ, which includes four components (as illustrated in Figure~\ref{fig:overview}): the construction of 1,375 OOD prediction tasks from six datasets, an uncertainty-aware training method based on Dropout and DER, evaluation metrics for both accuracy and uncertainty, and 12 representative GNN models.

\subsection{Datasets and OOD tasks}
To rigorously assess model performance under distributional shifts, we construct 1,375 OOD prediction tasks derived from six curated datasets. These tasks are generated through six data-splitting strategies: five established OFM-based methods and a novel SOAP-based approach, detailed in the following.

\paragraph{Datasets.} 
We select six benchmark datasets spanning diverse materials domains. Five datasets—dielectric (D1), log\_gvrh (D2), perovskites (D3), mp\_gap (D4), and jdft2d (D5)—are sourced from the MatBench suite \cite{MatbenchDunn2020}, representing optical, mechanical, thermodynamic, electronic, and thermodynamic properties, respectively. In addition, we include SuperCon3D (D6) \cite{chen2024learning}, a dataset for predicting the superconducting transition temperature \cite{hansen2022superconductivity} of 3D materials \cite{bergerhoff1983inorganic}. Table~\ref{Data} summarizes the dataset statistics and property types.

\begin{table}[ht!]
\caption{Summary of datasets.}
\label{Data}
\begin{tabular*}{\textwidth}{@{\extracolsep{\fill}}lllll@{\extracolsep{\fill}}}
\toprule
\textbf{Dataset} & \textbf{Samples} & \textbf{Target property} & \textbf{Data Source} &\textbf{Property Type} \\ 
\midrule
D1    & 4,764            & Refractive index    & Materials Project \cite{jainmaterials, ong2015materials, petousis2017high} & Optical \\
D2      & 10,987           & Shear modulus            & Materials Project \cite{jainmaterials, ong2015materials} & Mechanical   \\
D3    & 18,928           & Formation energy         & Castelli et al.\cite{castelli2012new} & Thermodynamic \\
D4    & 106,113      & Band gap         &  Materials Project \cite{jainmaterials, ong2015materials} & Electronic \\
D5    & 636           &  Exfoliation energy         &  JARVIS DFT 2D \cite{choudhary2017high}  & Thermodynamic \\
D6   &  1,016          & Transition temperature         & Chen et al. \cite{chen2024learning} &  Thermal \\ \bottomrule
\end{tabular*}
\end{table}

\paragraph{OOD tasks generation.}
To simulate distributional shifts across structure and property domains, we construct OOD test sets using five OFM-based and one SOAP-based splitting strategy. These strategies are designed to probe model generalization in low-density, rare, or structurally distinct material spaces.
\textbf{OFM-based OOD generation.}

The OFM encodes crystal structures, capturing structural similarities. We use five OFM-based splitting strategies~\cite{LOCO,SparseXY}: LOCO clusters 1024-dimensional OFM features into 50 OOD test sets; SparseXsingle(SXS) selects 50 low-density 2D OFM representatives; SparseXcluster(SXC) includes their 10 neighbors; SparseYsingle(SYS) and SparseYcluster(SYC) use property values for density.
{\bf SOAP-based OOD generation.}

To complement OFM-based splitting methods, we introduce SOAP-LOCO, a SOAP-based approach that captures fine-grained local atomic environments for OOD task generation. Using the DScribe library \cite{himanen2020dscribe}, we compute atomic-level SOAP descriptors, with dimensionality determined by the dataset’s maximum element types, radial basis functions \cite{choudhary2021atomistic}, and spherical harmonics order. These descriptors encode local geometric arrangements around each atom. Material-level representations are derived by aggregating atomic descriptors by element type into a matrix, with rows corresponding to element types, then flattening this matrix into a single vector.
SOAP-LOCO: We apply \( k \)-means clustering to these representations, treating each cluster as an OOD test set, with remaining samples split into training and validation sets at a 4:1 ratio. Unlike OFM-based splits with fixed 50 clusters, we dynamically optimize cluster counts for D1-D3 and D5-D6 to maximize MAE across folds, enhancing computational efficiency and generating challenging OOD tasks, as higher MAE reflects poorer model generalization. Optimized cluster counts for D1, D2, D3, D5, and D6 are 40, 20, 5, 40, and 10, respectively. For D4, high computational costs limit clusters to 5. Details on cluster optimization are provided in Appendix A.2.
In contrast to established OFM-based methods, SOAP-LOCO represents the first OOD test generation framework that capitalizes on SOAP's ability to capture detailed local atomic environments, providing a complementary and structurally informed perspective for materials property prediction evaluation.

In summary, the MatUQ benchmark yields 32 OOD scenarios, comprising 1,375 tasks. Each scenario corresponds to a dataset’s splitting strategy, with each cluster within a split defining one OOD task. Specifically, five datasets (D1-D3, D5-D6) each employ all six splits, contributing 1,365 tasks (1,250 from OFM-based splits at 50 tasks each, 115 from SOAP-LOCO). D4, constrained by computational costs, uses two splits (LOCO, SOAP-LOCO), adding 10 tasks (5 per split). These scenarios and tasks enable comprehensive model evaluation in Section 4.

\subsection{Uncertainty-aware training}
\label{subsec:uet}
Accurately quantifying uncertainty in OOD prediction is critical for trustworthy materials discovery. However, conventional training losses (e.g., MSE) only optimize prediction accuracy and do not encourage meaningful uncertainty calibration. We adopt dropout and DER~\cite{DeepEvidentialRegression} to address this gap: Dropout provides a Bayesian approximation that introduces stochasticity during inference, enabling epistemic uncertainty estimation, while DER directly models both epistemic and aleatoric uncertainty in a single forward pass via the Normal Inverse-Gamma (NIG) distribution. Together, they offer complementary strengths for robust uncertainty-aware training under distributional shifts.

\paragraph{Dropout regularization.}
We insert a spatial dropout layer (rate = 0.1) before the final output layer. This stochastic deactivation of neurons during training helps prevent overfitting and enables Monte Carlo-style uncertainty estimation at inference time, without modifying the model architecture.
\paragraph{Evidential loss.} 
Given a dataset $\mathcal{D} = \{(\boldsymbol{x}_i, y_i)\}_{i=1}^N$ of $N$ paired training examples, we aim to learn a functional mapping $f$ parameterized by weights $\boldsymbol{w}$ that approximately solves the optimization problem:
\begin{equation*}
\min_{\boldsymbol{w}} \mathcal{L}^{\text{DER}}(\boldsymbol{w}), \quad \mathcal{L}^{\text{DER}}(\boldsymbol{w}) = \frac{1}{N} \sum_{i=1}^N \left( \mathcal{L}_i^{\text{NLL}}(\boldsymbol{w}) + \lambda \mathcal{L}_i^{\mathrm{R}}(\boldsymbol{w}) \right),
\end{equation*}
where the regularization weight $\lambda$ balances prediction accuracy and uncertainty calibration. The negative log-likelihood (NLL) loss $\mathcal{L}_i^{\text{NLL}}(\boldsymbol{w})$ quantifies the model’s confidence in its predictions:
\[
\mathcal{L}_i^{\text{NLL}}(\boldsymbol{w}) = \frac{1}{2} \log ( \frac{\pi}{\nu_i} ) - \alpha_i \log(\Omega) + ( \alpha_i + \frac{1}{2} ) \log \left( (y_i - \gamma_i)^2 \nu_i + \Omega \right) + \log  \frac{\Gamma(\alpha_i)}{\Gamma(\alpha_i + \frac{1}{2})},
\]
and the regularization term $\mathcal{L}_i^{\mathrm{R}}(\boldsymbol{w})$ penalizes large prediction errors:
\begin{equation*}
\mathcal{L}_i^{\mathrm{R}}(\boldsymbol{w}) = |y_i - \gamma_i| (2 \nu_i + \alpha_i).
\end{equation*}
Here, $\Omega = 2 \beta_i (1 + \nu_i)$, $\Gamma(\cdot)$ denotes the gamma function, and $(\gamma_i, \nu_i, \alpha_i, \beta_i)=f(\boldsymbol{x}_i;\boldsymbol{w})$ are the evidential distribution parameters output by the neural network for each target $y_i$.
By jointly training with DER and dropout, the model learns both accurate predictions and well-calibrated uncertainty, especially under challenging OOD settings.

\subsection{Uncertainty-aware metrics}

To evaluate the accuracy and reliability of OOD predictions, we propose a robust protocol assessing models under deterministic and stochastic inference settings. Deterministic inference uses standard predictions, while stochastic inference employs MCD \cite{MCdropout2017uncertainties} with \( T \) forward passes to capture model uncertainty. Our protocol includes five metrics: two for deterministic inference (MAE, EviU) and three for stochastic inference (D-MAE, D-Unc, D-EviU), with the novel D-EviU metric as the primary evaluation metric for comprehensive uncertainty quantification, demonstrated by its superior performance in uncertainty assessments in Section 4.

We first introduce two metrics computed using deterministic inference. (1) Mean Absolute Error (MAE) measures predictive accuracy as the average absolute difference between predictions \( \hat{y}_i \) and ground truth \( y_i \) across \( N \) test samples: \( \text{MAE} = \frac{1}{N} \sum_{i=1}^{N} |\hat{y}_i - y_i| \).
(2)
Evidential Uncertainty (EviU) is derived from DER. 
For each sample \( i \), EviU uses NIG hyperparameters \( (\gamma_i, \nu_i, \alpha_i, \beta_i ) \) to compute: \( \text{EviU} = \frac{1}{N} \sum_{i=1}^{N} \left( \frac{\beta_i}{\nu_i (\alpha_i - 1)} + \frac{\beta_i}{\alpha_i - 1} \right) \).

 To evaluate the impact of stochastic inference, we use MCD with \( T = 50 \) forward passes and introduce two additional metrics: (1) Dropout-enhanced MAE (D-MAE), 
assessing accuracy under stochastic inference, and (2) Dropout Uncertainty (D-Unc), which estimates epistemic uncertainty via prediction variance. However, while D-Unc effectively captures model uncertainty, it does not account for inherent data noise. To provide a more complete characterization of uncertainty, we further incorporate (3) Dropout-enhanced EviU (D-EviU), quantifies both epistemic (model-driven) and aleatoric (data-driven) uncertainties. D-EviU extends EviU by averaging uncertainty estimates across stochastic passes for a robust total uncertainty measure. 

Below are detailed definitions of these three metrics. D-MAE extends MAE by averaging \( T \) MCD forward-pass predictions \( \gamma_{i,t}  (1\leq t\leq T) \) for each sample \( i \) to yield its predicted value, then applying the MAE formula to compute the error.
D-Unc quantifies epistemic uncertainty by measuring the variance of predicted means across the \( T \) dropout passes:
\( \text{D-Unc}_i = \frac{1}{T} \sum_{t=1}^{T} (\gamma_{i,t} - \hat{y}_i)^2, \)
where \( \gamma_{i,t} \) is the predicted value in the \(t\)-th pass and \(\hat{y}_i\) is the sample’s average prediction. 
D-EviU extends EviU by computing evidential uncertainty through the average parameters of the NIG distribution derived from \( T \) stochastic forward passes. Specifically, for each sample \( i \) and pass \( i \), the parameters \(\nu_{i,t}\), \(\alpha_{i,t}\) and \(\beta_{i,t}\) are averaged over \( T \) passes to form the final NIG parameters, which are then used to calculate the evidential uncertainty as defined in EviU.

\subsection{Baseline GNN models}
We evaluate 12 representative GNN models published between 2018 and 2025, covering both convolution-based and attention-based architectures. These models span the evolution of GNN design in materials property prediction and reflect diverse inductive biases and message-passing mechanisms.
\paragraph{Convolution-based architectures.}
The early generation of GNNs for crystalline materials relies on convolutional operations to capture local chemical environments.
CGCNN \cite{xie2018crystal} introduced a basic graph convolutional architecture that aggregates neighboring atomic features, with interatomic distances encoded via radial basis functions (RBF).
SchNet \cite{schutt2018schnet} replaced discrete filters with continuous ones and used a cosine cutoff function to ensure smooth distance-based interactions.
MEGNet \cite{chen2019graph} augmented the message-passing process by integrating global state features, improving the model's ability to capture system-level dependencies.
DimeNet++ \cite{gasteiger2020fast} explicitly modeled angular relationships and three-body interactions using spherical Bessel functions and a two-step edge-based message passing scheme.
ALIGNN \cite{choudhary2021atomistic} extended message passing to line graphs, incorporating bond-angle information via edge-to-edge, edge-to-node, and node-to-node interactions.
PotNet \cite{lin2023efficient} embedded physics-based long-range interactions by modeling interatomic potentials (Coulomb, dispersion, repulsion) through infinite summation in the message passing process.
eComFormer \cite{yan2024complete} ensures SO(3)-equivariance using spherical harmonic-based vector embeddings.

\paragraph{Attention-based architectures.}
Recent efforts have shifted toward attention-based models that capture complex dependencies across atomic graphs.
DeeperGATGNN \cite{omee2022scalable} stabilized deep attention layers through group normalization and residual connections, enabling scalable training of deeper GNNs.
Matformer \cite{yan2022periodic} fused graph convolution with self-attention, incorporating periodicity-aware representations that preserve lattice structure invariance.
iComFormer \cite{yan2024complete} encodes SE(3)-invariant geometric interactions via three-body features.
SODNet \cite{chen2024learning} further generalized geometric representation learning by handling both ordered and disordered crystals with spherical harmonics and type-weighted vector attention mechanisms.
CrystalFramer \cite{ito2025crystalframer} introduced dynamic frames that leverage attention mechanisms to align coordinate systems with learned interatomic interactions while maintaining SE(3) invariance.

For simplicity, we will refer to the twelve models-CGCNN, SchNet, MEGNet, DimeNet++, ALIGNN, PotNet, eComFormer, DeeperGATGNN, Matformer, iComFormer, SODNet, and CrystalFramer-as M1 through M12, respectively, in the following sections. Detailed descriptions of these models, including their associated hyperparameters, are provided in the Appendix B.
\section{Experiments on uncertainty-aware OOD evaluation}
\label{sec:Experiments}
To evaluate 12 GNN models (Section 3.4), we train them using uncertainty-aware methods (Section 3.2) on 1,375 OOD tasks across 32 scenarios derived from six datasets (Section 3.1), assessed with five performance metrics (Section 3.3). Experiments on D1-D3 are conducted on NVIDIA RTX 4090 GPUs (CUDA 11.8) with PyTorch 2.1.2 and TensorFlow 2.9.0, while experiments on D4-D6 are performed on NVIDIA H100 GPUs (CUDA 12.1) with PyTorch 2.4.0 and TensorFlow 2.18.0.
This section presents our evaluation framework and experimental results. Subsection 4.1 justifies our novel data-splitting and training strategies, while Subsection 4.2 highlights GNN performance for each dataset’s most challenging OOD scenario, using optimal uncertainty metrics. Detailed experimental settings and comprehensive results are provided in Appendices C and D, respectively.

\subsection{Validating the MatUQ framework}
\label{sec:exp}
\paragraph{Novel SOAP-LOCO OOD generation.}
Table \ref{tab:performance_reorganized} presents the challenge levels of OOD splitting strategies, measured by average MAE across 12 models.
The results show the proposed SOAP-LOCO method generates the highest MAE for most datasets (D2-D6), with only SXS producing higher MAE for D1. This confirms SOAP-LOCO creates exceptionally challenging OOD scenarios, which is expected since GNNs represent crystals structurally and SOAP is a structural descriptor, naturally creating structural OOD conditions that rigorously test model generalization capabilities.
\begin{table}[!h]
\centering
\caption{
MAE-based OOD challenge levels for OOD splitting strategies across six datasets. Higher MAE reflects the severity of OOD challenge and corresponding difficulty in model generalization. \textbf{Bold} indicates the most demanding OOD scenarios per dataset.
}
\begin{tabular*}{\textwidth}{@{\extracolsep{\fill}}lcccccc@{\extracolsep{\fill}}}
\toprule
{Dataset} &  LOCO & SXC & SYC & SXS & SYS & SOAP-LOCO \\
\midrule
D1 & 0.544 & 0.572 & 0.369 & \textbf{1.572} & 0.218 & 0.300 \\
D2 & 0.142 & 0.116 & 0.092 & 0.118 & 0.071 & \textbf{0.159} \\
D3 & 0.134 & 0.101 & 0.088 & 0.063 & 0.037 & \textbf{0.158} \\
D4 & 0.486 & - & - & - & - & \textbf{0.669} \\
D5 & 46 & 48 & 44 & 59 & 33 & \textbf{65} \\
D6 & 14.067 & 3.806 & 3.802 & 6.991 & 2.610 & \textbf{76.390} \\
\bottomrule
\end{tabular*}
\label{tab:performance_reorganized}
\end{table}

\paragraph{Improving prediction accuracy via uncertainty-aware training.} 
Table~\ref{tab:trainingvs} compares the MAE of our uncertainty-aware MatUQ framework with that of standard MSE loss training. 
The results show that MatUQ achieves an average MAE reduction of 70.61\% across datasets D1-D3, with a remarkable 84.49\% reduction on D3, substantially enhancing model prediction accuracy.
This improvement stems from two synergistic mechanisms: evidential loss ensures robust uncertainty calibration, while dropout regularization mitigates overfitting, enhancing generalization under distribution shifts.
\begin{table}[!h]
\centering
\begin{minipage}{0.51\textwidth}
    \centering
        \caption{MAE of uncertainty-aware MatUQ vs. standard MSE across D1-D3. 
         MAE is averaged over five OFM-based OOD splits and seven models (M1-M5, M8, M12). Standard training results are from~\cite{omee2022scalable}, except for M12 (implemented by us).
         \textbf{Bold} indicates the best results.
        }
    \label{tab:trainingvs}
\begin{tabular}{lccc} 
\toprule
\multirow{2}{*}{Dataset} & \multicolumn{2}{c}{Training method} &\multirow{2}{*}{Reduction (\%)} \\
\cmidrule{2-3}
& Standard & MatUQ &  \\
\midrule
D1 & 1.695 & \textbf{0.674} & 60.26 \\
D2 & 0.658 & \textbf{0.110} & 83.30 \\
D3 & 0.662 & \textbf{0.103} & 84.49 \\
\midrule
Avg & 1.005 & \textbf{0.295} & 70.61 \\
\bottomrule
\end{tabular}
\end{minipage}
\hfill
\begin{minipage}{0.46\textwidth}
    \centering
     \caption{
     Spearman correlations between uncertainty metrics and prediction errors. \textbf{Bold} indicates the highest correlation values, representing more effective UQ methods for each dataset.
     }
    \label{tab:correlation_coefficients}
    \begin{tabular}{lccc}
        \toprule
        \multirow{2}{*}{Dataset} & \multicolumn{3}{c}{Spearman correlations} \\
        \cmidrule{2-4}
          & EviU & D-Unc & D-EviU \\
        \midrule
        D1 & \textbf{0.6114} & 0.3593 & 0.4295 \\
         D2 & 0.1117 & 0.0182 & \textbf{0.2535} \\
         D3 & 0.2170 & 0.1248  & \textbf{0.2171} \\
        D4 & 0.1821 & 0.1053 & \textbf{0.3188} \\
        D5 & 0.2936  & \textbf{0.4028}  & 0.3348 \\
        D6 & 0.0337  & 0.0730  & \textbf{0.1536} \\
        \bottomrule
    \end{tabular}
\end{minipage}
\end{table}

\subsection{Benchmarking model performance under distribution shifts}

We present representative performance results for 12 GNN models on high-difficulty OOD scenarios across six datasets in Table~\ref{tab:performance}, with metrics averaged over all OOD test tasks within each scenario. 
The statistically significant OOD scenarios for each dataset, specifically SOAP-LOCO for D2–D6 and SXS for D1, demonstrate substantial distribution shifts as evidenced in Table~\ref{tab:performance_reorganized}.
For each scenario, we report two carefully selected metrics due to space constraints, an optimal uncertainty metric (EviU for D1, D-EviU for D2-D4 and D6, D-Unc for D5) determined by the highest Spearman rank correlation coefficients \cite{spearman1961proof} from Table~\ref{tab:correlation_coefficients}, and a corresponding accuracy metric (D-MAE for dropout-based uncertainty metrics, or MAE otherwise) selected accordingly.
The Spearman correlation coefficients in Table~\ref{tab:correlation_coefficients}, averaged across all models, quantitatively evaluate the alignment between uncertainty estimates and actual prediction errors \cite{gneiting2007strictly}. Higher correlation values indicate more reliable UQ, enabling the selection of the most effective uncertainty metric for each OOD scenario.

\begin{table}[!h]
\centering
\caption{Performance of 12 GNN Models on Challenging OOD Scenarios. \textbf{Bold} indicates the best results, \underline{underline} the second best.
}
\label{tab:performance}
\scriptsize  
\setlength{\tabcolsep}{2.5pt}  
\begin{tabular*}{\textwidth}{@{\extracolsep{\fill}}lcccccccccccc@{\extracolsep{\fill}}}
\toprule
\multirow{2}{*}{\textbf{Model}} & \multicolumn{2}{c}{\textbf{D1}} & \multicolumn{2}{c}{\textbf{D2}} & \multicolumn{2}{c}{\textbf{D3}} & \multicolumn{2}{c}{\textbf{\textbf{D4}}} & \multicolumn{2}{c}{\textbf{\textbf{D5}}} & \multicolumn{2}{c}{\textbf{\textbf{D6}}}\\
\cmidrule{2-3}\cmidrule{4-5}\cmidrule{6-7}\cmidrule{8-9}\cmidrule{10-11}\cmidrule{12-13}
& \textbf{MAE} & \textbf{EviU} 
& \textbf{D-MAE} & \textbf{D-EviU} 
& \textbf{D-MAE} & \textbf{D-EviU} 
& \textbf{D-MAE} & \textbf{D-EviU} & \textbf{D-MAE} & \textbf{D-Unc} & \textbf{D-MAE} & \textbf{D-EviU}\\
\midrule
M1[2018] & 1.547 & 3.3\text{e+5} 
& 0.359 & \underline{6.6e-2} 
& 0.175 & 9.1\text{e-3} 
& 0.863 & 5.6e+2 
& 77.7        & 1.5e-3
& 146   & 5.5e-1
\\
M2[2018] & 1.573 & 1.1\text{e+2}
& \underline{0.128} & 2.3e-1 
& \underline{0.061} & 7.3\text{e-3}
& 0.575 & 3.0e+0 
& \underline{63.0} & 2.7\text{e-3} 
& 69.5e+6         & 2.7e+16 
\\
M3[2019] & 1.674 & 1.3\text{e+5}
& 0.145 & 2.6e-1 
& 0.071 & 1.4\text{e-2} 
& 0.874 & 2.5e+0 
& 66.3        & 4.0\text{e-3} 
& 82    & 6.0e+0
\\
M4[2020] & 1.583 & 3.6\text{e+4}
& 0.213 & 1.5e-1 
& 0.729 &  5.0\text{e-1} 
& 0.444     & 2.0\text{e+5}     
& 67.4        & \underline{8.7\text{e-4}}
& 28.7e+6         & 1.7e+15  
\\
M5[2021] & 1.551 & 2.3\text{e+3} 
& 0.186 & 1.7e-1 
& \textbf{0.046} & \underline{4.3\text{e-3}} 
& \textbf{0.369} & \textbf{1.0e-1}
& 67.4        & 3.9\text{e-2}
& \textbf{61}    & 4.3e+0
\\
M6[2023] & 1.558 & 1.5\text{e+1} 
& 0.366 & 8.5e-2 
& 0.138 & 4.6\text{e-3}
& 0.827 & \underline{1.5e-1}
& 81.4        & 4.0\text{e-3}
& 129   & 1.1e-1
\\
M7[2024] & 1.535 & 1.4\text{e+1}
& 0.351 & 8.1e-2 
& 0.134 & 6.5\text{e-3} 
& 0.793 & 2.1e+0
& 77.9        & 5.3\text{e-3}
& 129   & \underline{5.2e-2} 
\\
\addlinespace
M8[2022] & 1.642 &5.4\text{e+4} 
& 0.279 & 2.0e-1 
& 0.074 & \textbf{4.0\text{e-3}} 
& 0.763 & 1.1e+3
& 69.4        & \textbf{6.4\text{e-4}}
& 123   & 4.5e+4
\\
M9[2022] & 1.587 & \underline{1.2\text{e+0}} 
& 0.377 & 8.5e-2 
& 0.162 & 5.9\text{e-3}
& 0.760 & 2.5e-1
& 85.0        & 4.2\text{e-3}
& 128   & 6.2e-2\\
M10[2024] & \underline{1.515} & 7.8\text{e+2} 
& 0.352 & \textbf{5.7e-2} 
& 0.138 &  4.8\text{e-3}
& 0.761 & 2.8e+0
& 76.3        & 5.8\text{e-3}
& 130   & \textbf{1.1e-2}
\\
M11[2024] & 1.596 & \textbf{9.1\text{e-1}}
& 0.145 & 1.7e-1 
&  0.074 &  5.4\text{e-3}
& 0.523 & 1.7e+0
& \textbf{62.6}        & 9.0\text{e-4}
& 97    & 2.0e+1
\\
M12[2025] & \textbf{1.507} & 1.2\text{e+2} 
& \textbf{0.096} & 8.1e-2 
& 0.077 & 4.7\text{e-3} 
& \underline{0.427}     & 3.0\text{e-1}   
& 67.1        & 3.3\text{e-3}
& \underline{71}   & 1.2e+1
\\
\bottomrule
\end{tabular*}
\end{table}

\subsubsection{Model-specific insights}
\paragraph{Early convolutional architectures.}
Our analysis of early GNN models reveals their strengths and limitations in materials property prediction. CGCNN performs poorly overall, achieving a moderate mean absolute error (MAE) of 1.547 on the D1 dielectric dataset but with extremely high uncertainty (3.3e+5), indicating its inability to model complex properties requiring local and long-range interactions. SchNet ranks second in precision on D2, D3, and D5, leveraging continuous filter functions to capture short-range interactions, but it is unreliable on D2 and D3, with moderate uncertainty on D5 and catastrophic failure on D6 (MAE 69.5e+6, uncertainty 2.7e+16) due to missing quantum many-body effects. MEGNet incorporates global features, achieving balanced precision but poor uncertainty quantification (UQ), as global descriptors without proper structural encoding prioritize precision over reliability. DimeNet++ shows inconsistent performance, with strong precision on D4 but the worst uncertainty, and catastrophic results on D3 and D6, highlighting how overemphasizing directional features undermines reliability for isotropic and quantum properties.

\paragraph{Advanced convolutional approaches.}
ALIGNN excels on D3 and D4, leveraging its dual-graph architecture to capture connectivity and bond angles, making it highly effective for coordination-dependent properties like formation energy and bandgaps. PotNet and eComFormer achieve moderate performance across datasets. PotNet’s interatomic potential summations focus on local structural features, while eComFormer’s vector representations within convolutional frameworks struggle to balance short- and long-range interactions. DeeperGATGNN provides excellent uncertainty quantification (UQ) on D3 and D5 through attention mechanisms but underperforms on electronic properties requiring quantum modeling.

\paragraph{Transformer-based architectures.}
Matformer demonstrates reliable SE(3) and periodic invariance encoding on D1 but only moderate overall precision, highlighting that effective periodicity modeling requires specific interaction mechanisms, particularly for D5's weak interlayer forces.
iComFormer advances periodic invariance formulations with best MAE on D1 (1.515) and excellent uncertainty on D2 and D6, effectively capturing periodic patterns for optical and mechanical properties while potentially under-modeling local electronic features.
SODNet performs inconsistently, excelling on D5 through its transformer-based vector SE(3) equivariance for layered materials' anisotropy, yet struggling elsewhere due to imbalanced local and long-range interactions interaction modeling, while its poor D1 precision contrasts with best-in-class uncertainty metrics suggesting unexplored reliability potential.
CrystalFramer (M12) achieves remarkable precision across six datasets through dynamic reference frames that adaptively focus on relevant coordination environments, effectively balancing local precision with global context.

\paragraph{Technical evolution trends.}
Our analysis identifies four key trends in materials property prediction. First, geometric expressivity has evolved from simple distance-based encodings to angular feature integration and dynamic coordinate systems. Structure-aware architectures, such as CrystalFramer, ALIGNN, and SODNet, provide richer geometric representations, improving prediction accuracy and domain transferability. Second, computational paradigms have shifted from early graph neural network (GNN) message-passing limitations to transformer-based frameworks (M9-M12), enhancing uncertainty quantification (UQ) critical for reliable material screening. Third, physics-informed constraints remain essential, as demonstrated by SchNet’s strong performance on mechanical properties despite its 2018 origin, reflecting the effectiveness of continuous filter functions for modeling physical interactions. Fourth, appropriate inductive biases improve generalization, reducing dielectric prediction uncertainty from e+5 to e-1, enabling exploration of novel material design spaces. These trends highlight structure-awareness as key to reliable predictions, suggesting future architectures should balance physical constraints with representational flexibility to achieve precise predictions and robust UQ across diverse material properties.

\subsubsection{Dataset-specific insights}
The six datasets span diverse material properties, each presenting unique challenges that influence model performance. Analysis of the dielectric dataset reveals that transformer architectures deliver substantially improved uncertainty (9.1e-1 to 7.8e+2) compared to convolutional models (1.1e+2 to 3.3e+5), aligning with dielectric response physics requiring both local coordination and long-range ordering. For log\_gvrh elastic properties, CrystalFramer excels (MAE 0.096) through dynamic reference frames, while iComFormer delivers the best uncertainty (5.7e-2) through periodic invariance on the lattice. SchNet's effective modeling of short-range forces yields the second-best precision (0.128) on this dataset. The perovskites dataset highlights ALIGNN's outstanding performance (MAE 0.046) through angular information modeling, while DeeperGATGNN provides the most reliable uncertainty (4.0e-3) through attention mechanisms that effectively recognize perovskite coordination patterns. Similarly, for mp\_gap electronic property prediction, ALIGNN achieves exceptional results (MAE 0.369) by capturing orbital hybridization through bond angles, while CrystalFramer and SODNet effectively model coordination environments where band states localize. For 2D materials in jdft2d, SODNet delivers optimal performance (MAE 62.6) through vector equivariance that captures directional anisotropy in layered materials, while SchNet models van der Waals interactions effectively. The SuperCon3D dataset presents the greatest challenge, with GNN models exhibiting significant performance variations. SchNet and DimeNet++ fail catastrophically on superconductivity prediction, while ALIGNN and CrystalFramer excel through angular information and dynamic frames capable of capturing electron-phonon coupling. Transformer-based models provide excellent UQ despite moderate precision on this complex quantum property.
\section{Conclusions}
\label{sec:Conclusions}
We introduce MatUQ, a comprehensive benchmark framework for evaluating uncertainty-aware GNNs in OOD materials property prediction. Our SOAP-LOCO data splitting strategy creates challenging test scenarios by capturing fine-grained local atomic environments. The uncertainty-aware training approach, combining MCD and DER, reduces prediction errors by up to 70.61\%. The proposed D-MAE and D-EviU metrics provide a comprehensive evaluation framework capturing both predictive accuracy and UQ through stochastic inference, with D-EviU showing superior correlation with prediction errors on four of six benchmarks. 
Comparative analysis of twelve representative architectures reveals that geometric-aware transformers (CrystalFramer, ALIGNN, SODNet) consistently outperform conventional GNNs across diverse properties. Our findings establish optimal model selections for specific applications: SODNet for dielectric, CrystalFramer for elastic properties, ALIGNN for perovskite design, superconductivity screening and bandgap engineering (with CrystalFramer as an alternative), and SODNet for 2D material exfoliation energies.

Despite these advances, challenges persist. Dataset characteristics, beyond sample size, significantly affect model performance. Quantum mechanical properties, such as superconductivity and bandgap, are harder to predict than classical thermodynamic or mechanical properties like perovskite or elastic properties. Moving forward, we believe that incorporating lessons from these findings—particularly the importance of UQ and structure-aware representations—could contribute to more reliable materials prediction models. By publicly releasing MatUQ, we hope to provide a useful resource that emphasizes the critical importance of uncertainty-aware training and evaluation for robust predictions in real-world materials discovery, where distribution shifts are inevitable. Our future work will focus on extending these methods to additional materials domains and exploring more sophisticated uncertainty calibration techniques to further address the challenges identified in this study.

\medskip

{
\small

\bibliographystyle{plain}
\bibliography{ref}
}

\end{document}